\def\year{2022}\relax
\documentclass[letterpaper]{article} % DO NOT CHANGE THIS
\usepackage{aaai22}  % DO NOT CHANGE THIS
\usepackage{times}  % DO NOT CHANGE THIS
\usepackage{helvet}  % DO NOT CHANGE THIS
\usepackage{courier}  % DO NOT CHANGE THIS
\usepackage[hyphens]{url}  % DO NOT CHANGE THIS
\usepackage{graphicx} % DO NOT CHANGE THIS
\usepackage{natbib}  % DO NOT CHANGE THIS AND DO NOT ADD ANY OPTIONS TO IT
\usepackage{caption} % DO NOT CHANGE THIS AND DO NOT ADD ANY OPTIONS TO IT
\DeclareCaptionStyle{ruled}{labelfont=normalfont,labelsep=colon,strut=off} % DO NOT CHANGE THIS
\usepackage{bm}
\usepackage{amssymb}
\usepackage{multirow}
\usepackage{graphicx}
\usepackage{tabularx}
\usepackage[acronym]{glossaries}
\usepackage{makecell}
\usepackage{rotating}
\usepackage{threeparttable}
\usepackage{booktabs}
\usepackage{ragged2e}
\usepackage{array}

\usepackage{color}
\usepackage{transparent}
\usepackage{graphicx}
\usepackage{import}

\newcommand{\executeiffilenewer}[3]{%
 \ifnum\pdfstrcmp{\pdffilemoddate{#1}}%
 {\pdffilemoddate{#2}}>0%
 {\immediate\write18{#3}}\fi%
}

\newcommand{\ie}{\textit{i.e.}, }
\newcommand{\eg}{\textit{e.g.}, }

\usepackage{algorithm}
\usepackage{algorithmic}

\usepackage{newfloat}
\usepackage{listings}
\floatstyle{ruled}
\newfloat{listing}{tb}{lst}{}
\floatname{listing}{Listing}
\title{Rethinking Two Consensuses of Transferability in Deep Learning}
\author{
    Yixiong Chen \textsuperscript{\rm 1},
    Jingxian Li \textsuperscript{\rm 2},
    Chris Ding \textsuperscript{\rm 1},
    Li Liu \thanks{Corresponding author: liuli@cuhk.edu.cn} \textsuperscript{\rm 1,3}
}
\affiliations{
    \textsuperscript{\rm 1} The Chinese University of Hong Kong - Shenzhen, Shenzhen, China\\
    \textsuperscript{\rm 2} Fudan University, Shanghai, China\\
    \textsuperscript{\rm 3} Shenzhen Research Institute of Big Data, Shenzhen, China\\
}

\usepackage{bibentry}
\begin{document}

\maketitle

\begin{abstract}
Deep transfer learning (DTL) has formed a long-term quest toward enabling deep neural networks (DNNs) to reuse historical experiences as efficiently as humans. This ability is named \textit{knowledge transferability}. A commonly used paradigm for DTL is firstly learning general knowledge (pre-training) and then reusing (fine-tuning) them for a specific target task. There are two consensuses of transferability of pre-trained DNNs: (1) a larger domain gap between pre-training and downstream data brings lower transferability; (2) the transferability gradually decreases from lower layers (near input) to higher layers (near output).
However, these consensuses were basically drawn from the experiments based on natural images, which limits their scope of application. This work aims to study and complement them from a broader perspective by proposing a method to measure the transferability of pre-trained DNN parameters. Our experiments on twelve diverse image classification datasets get similar conclusions to the previous consensuses. More importantly, two new findings are presented, \textit{i.e.}, (1) in addition to the domain gap, a larger data amount and huge dataset diversity of downstream target task also prohibit the transferability; (2) although the lower layers learn basic image features, they are usually not the most transferable layers due to their domain sensitivity.

\end{abstract}

\section{Introduction}
Deep transfer learning (DTL) enables deep neural networks (DNNs) to learn faster and gain better generalization with limited data. This successful bionic design came from the knowledge \textit{transferability} of human brains \citep{ausubel1968educational}, which are capable of learning new concepts based on the original cognitive structure. DNNs transfer knowledge in a similar way. They learn general recognition abilities from large pre-training datasets (\eg ImageNet \citep{deng2009imagenet}) and fine-tune the network parameters on specific downstream datasets \citep{tajbakhsh2016convolutional}.

\begin{figure}[t]
\centering\centerline{\includegraphics[width=0.8\linewidth]{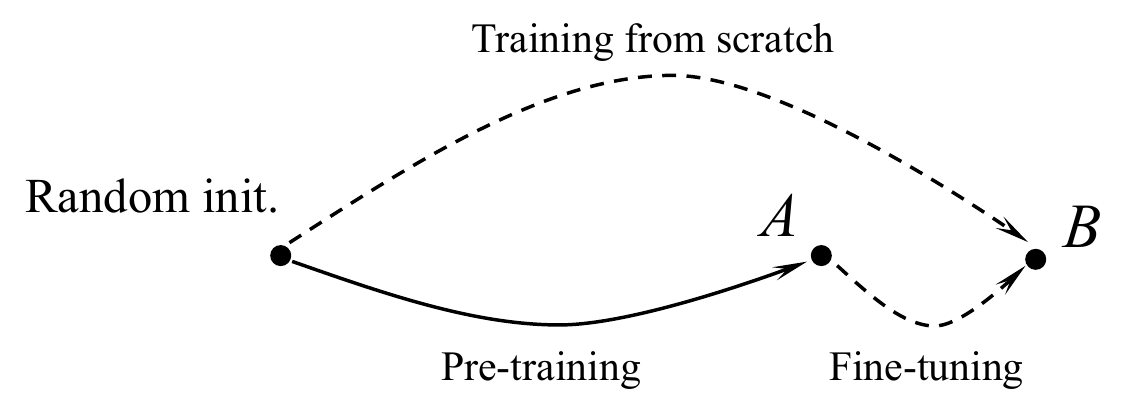}}
\caption{The motivation of our transferability measurement. Pre-training on task A makes the fine-tuning only needs shorter optimization distance to reach the solution of task B than training from scratch.}
\label{fig:Motivation}
\end{figure}

Previous literature \citep{yosinski2014transferable,fayek2018transferability,tajbakhsh2016convolutional} used semi-quantitative methods (\ie usually by comparing the network performance with/without transfer learning) to judge the DNNs' transferabilities. Two consensuses were drawn from these works. 
Firstly, the transferability decreases as the distance between the source and target dataset (domain gap) increases. We regard its root cause to be the poor transferability of DNNs to the samples dissimilar to pre-training data, which stimulates our interest in exploring other important factors leading to a greater extent of dissimilarity (e.g., dataset diversity). 
Secondly, the lowest layer (\ie the layer that is close to the input) learns general knowledge like colors and edges, while the highest layer (\ie the layer that is close to the output) learns semantic knowledge specific to the task, indicating there is a decreasing trend of transferability from low to high layers. Although this conclusion is generally correct, we conjecture that the pre-trained features of lower layers might not be that applicable to the downstream tasks. If the local pattern distributions of data change drastically (\eg from natural images to cartoons or even medical data), the shallower local features might suffer more than the deeper semantic features. In these cases, the decreasing trend of layer-wise transferability may need further validations.

Instead of using semi-quantitative measuring methods, this work revisits these consensuses with a quantitative method for more precise results. We mainly study the following two questions:
\begin{itemize}
\item \textit{In addition to the domain gap, what other critical factors about datasets affect the transferability?}
\item \textit{Do the transferabilities for sequential layers of a pre-trained DNN always follow a downward trend?}
\end{itemize}

% To answer these two questions, we start with the nature of knowledge transfer and propose a quantitative method for measuring transferability. Knowledge transfer is a procedure of reusing and adjusting learned abilities to do new tasks. For DNNs, pre-trained parameters can be regarded as learned knowledge. Better pre-trained parameters need less adjustment to gain power for downstream tasks. From the perspective of optimization, pre-training offers a better parameter initialization than random initialization, so that a few gradient steps of fine-tuning can produce good results on a new task \citep{finn2017model}. On this basis, we quantify transferability as how close parameters pre-trained on task A to the optimal parameters on task B compared with the distance between randomly initialized parameters to task B (Fig. \ref{fig:Motivation}). If the fine-tuning phase needs a shorter optimization distance, the transferability of parameters pre-trained on A would be higher. This quantification method makes transferabilities between the pre-training dataset and different downstream tasks comparable. We will also be able to derive the transferabilities between different layers precisely.

To answer these two questions, we start with the nature of knowledge transfer, a procedure of reusing and adjusting learned abilities for new tasks, to design our measuring method. For DNNs, pre-trained parameters can be regarded as learned knowledge, and it offers a better parameter initialization than random initialization (Fig. \ref{fig:Motivation}). In this case, a few gradient steps of fine-tuning can produce good results on a new task \citep{finn2017model}. On this basis, we quantify transferability as how much the pre-training task A pushes the parameters towards the optimal point of the downstream task B. 
If the fine-tuning phase needs a shorter optimization distance than training from scratch, pre-training helps to make the downstream adaptation easier.
This quantification method enables us to compare the transferabilities between a pre-training dataset and different downstream tasks under the same standard. It also provides a way to derive the transferabilities of different layers precisely. With this measurement, the above questions can be answered with solid justifications.

For the under-explored factors that affect transferability, downstream data diversity (named domain width in this work) and data amount come to our mind for simple reasons: (1) When the domain width of a target dataset goes larger, there must exist more out-of-distribution samples that the DNN did not see in the pre-training phase, which impairs the transferability; (2) For data amount of the downstream task, it is shown by previous works \citep{he2019rethinking,newell2020useful} that the greater amount of data decreases the benefit brought by pre-training. In the experiments, our results verify that domain width and data amount negatively affect the transferability even to a greater extent than the domain gap. For the layer-wise transferability, this work finds that the lowest layers may be domain-sensitive, which makes the trends downward when only excluding the first few layers.

To conclude, the main contributions of this work are as follows:

\begin{enumerate}
\item A novel method for quantifying the transferability of DNN parameters is proposed. 
\item By revisiting the previous consensuses through rich experimental validations on twelve datasets, we find two important phenomena in DTL: (1) the target domain width and data amount highly affect the transferability, even more than the domain gap; (2) the transferabilities of the lowest layers are usually prohibited by the domain shift. For all of our experiments with ResNet50, the lowest layer is not the most transferable among all layers.
%\item[3.] We show that transferability of a DNN between two tasks can be a good indicator of the design of fine-tuning scheme. A simple improved method can both accelerate the training and boost the performance.
\end{enumerate}

\section{Related Work}

\subsection{Transferability of DNNs}

Transferability of DNNs is the ability to gain transferable knowledge from the source domain and reuse the knowledge to decrease the generalization error on the target domain, under the distribution shift or the task discrepancy \citep{jiang2022transferability}. A common method to assess the transferability of DNNs is comparing the performance between fine-tuning and training from scratch \citep{yosinski2014transferable,matsoukas2022makes}. Given a downstream task, more significant performance improvement means the DNN has better transferability. There are several key factors affecting how transferable the parameters of a DNN are. First, the performance of the DNN on the pre-training dataset is positively correlated to its transferability to downstream tasks \citep{kornblith2019better} in most cases. Second, the more similar the source and target task are, the more transferable the DNNs would be \citep{yosinski2014transferable,dwivedi2019representation}. Third, the transferabilities of parameters are fragile to the large learning rate \citep{ro2021autolr}. A drastic change of highly transferable parameters may destroy their original knowledge, making them less transferable. This work also tries to excavate the properties for DNNs' transferabilities, but from the perspective of the dataset.

\subsection{Pre-training}

To improve DNNs' transferabilities, many pre-training datasets and methods are proposed. The most dominant pre-training dataset for vision tasks is ImageNet \citep{deng2009imagenet} for its large-scale data with high-quality annotations. Earlier literature mainly pre-train models in a supervised way \citep{he2019rethinking} with categorical labels. 
However, researchers found that ImageNet cannot cover the need for all visual applications because the large domain gap between natural and special images (\eg medical images) leads to poor transferability. Pre-training datasets and methods for specific domains~\citep{sermanet2018time,chen2021uscl,zhang2022hico} began to emerge for this reason. This work tests the transferability for large domain gaps, and tries to provide some practical guides when pre-trained models on specific domains are unavailable. Other than training schemes, models themselves are crucial for good transferabilities. Pre-training with big models like convolutional neural networks (CNNs) with hundreds of layers \citep{huang2017densely} and Vision Transformers (ViTs) \citep{dosovitskiy2020image} are proved to bring more benefits to downstream tasks than smaller ones. The measurement proposed in this work also validates that larger pre-trained models have higher transferabilities (see the supplementary material).

\begin{figure*}[t]
\centering\centerline{\includegraphics[width=0.8\linewidth]{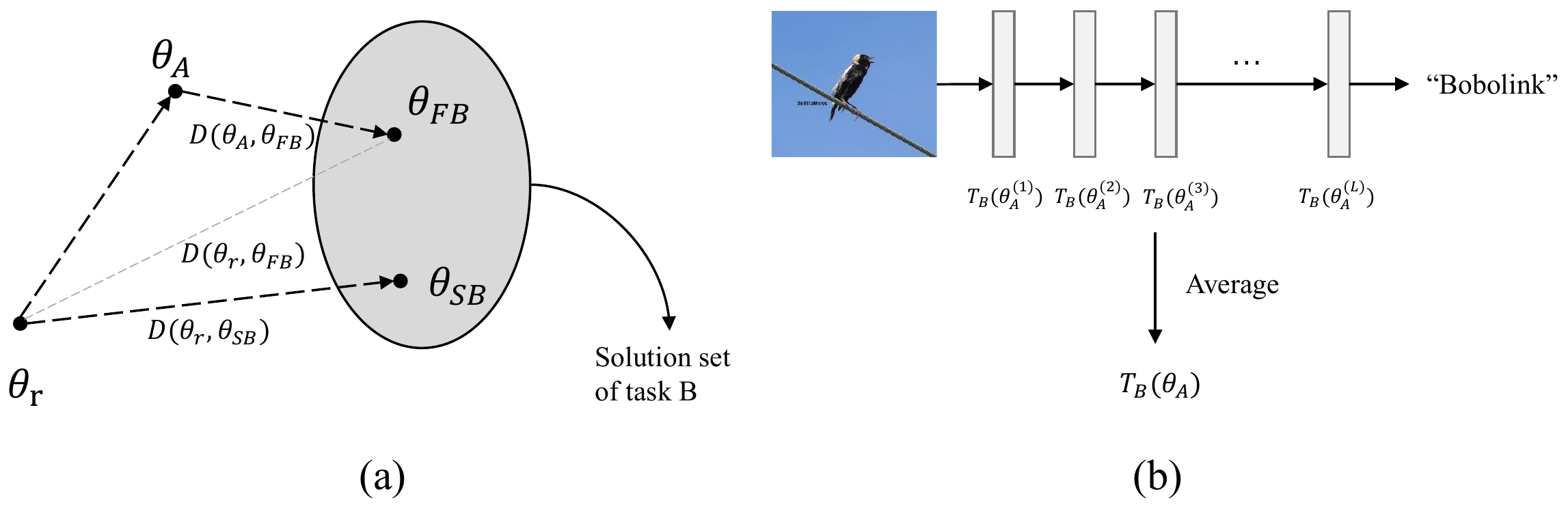}}
\caption{The method for measuring parameter transferability. (a) The transferability of the parameter $\theta_A$ is defined as $D(\theta_r, \theta_B)/D(\theta_A, \theta_B)$, but there is an ambiguity of $\theta_B$ between fine-tuning and training from scratch. We denote them as $\theta_{FB}$ and $\theta_{SB}$ respectively and use them to calculate two distance variants, $D(\theta_r, \theta_{FB})$ and $D(\theta_r, \theta_{SB})$. (b) The transferability of the whole network is calculated as the average transferability of each layer to avoid the parameter scale difference.}
\label{fig:Method}
\end{figure*}

\subsection{Fine-tuning}

How to leverage the transferabilities of pre-trained DNNs for a new task has become a research focus in recent years. The practices to transfer feature extraction capability from an off-the-shelf network to a new network can be roughly categorized into two classes: initialization-based and teacher-student-based \citep{song2021robust}. The former is the so-called fine-tuning method and is the default setting to measure transferabilities in this work. In addition to the common fine-tuning scheme that involves all layers, another powerful paradigm is tuning the last few layers of a pre-trained DNN \citep{tajbakhsh2016convolutional}, so that the low-level general features inherent in the shallow layers can be kept intact. Likewise, many advanced fine-tuning schemes \citep{long2015learning,li2019delta,zhong2020bi,ro2021autolr} assumed that the higher-level layers are more specific to the task, therefore, are less transferable and need more updates. However, recent research demonstrated some exceptions \citep{chen2022metalr} when domain shift is large, where lower layers also need to be adjusted to adapt to the different visual feature distributions. This finding is consistent with our experimental results, which reveal that the first few layers are not necessarily as transferable as researchers previously considered.

\section{Methodology}
To answer the above two questions, a general transferability measurement, domain-agnostic, task-agnostic, and architecture-agnostic, is proposed to compare parameter transferabilities considering different domains, tasks, and layers.

\paragraph{Definition of the Transferability.}
The aim for pre-training is to find a good initialization parameter $\theta_A$ through task A so that the optimal solution $\theta_B$ on task B can be obtained by fine-tuning as few steps as possible. We denote the distance between  $\theta_A$ and $\theta_B$ as $D(\theta_A, \theta_B)$, which reflects the transferability of $\theta_A$ to task B. However, since the optimization difficulties for different parameters in a DNN on different tasks are different, $D(\theta_A, \theta_B)$ cannot be regarded as transferability directly. To tackle this problem, the distance between random initialized parameters $\theta_{r}$ and $\theta_B$, $D(\theta_r, \theta_B)$, is utilized for normalization. This work defines the transferability of random initialized parameters to be 1, and measures the transferability of parameter $\theta_A$ to task B as follows:

\begin{equation}
    T_B(\theta_A) = \frac{D(\theta_r, \theta_B)}{D(\theta_A, \theta_B)}.
\end{equation}

When transferability is greater than 1, pre-training shortens the optimization distance, meaning that new knowledge is learned with the help of experiences. In practice, the distance metric can be calculated with mean absolute difference $D(\theta_A,\theta_B) = \sum_{i=1}^K |\theta_A^i - \theta_B^i|/K$, where $K$ is the number of parameters. 

\paragraph{Two Different Optimal Points.} One crucial problem for this measurement is, that in most cases, the pre-trained DNNs would not converge to the same (local) optima as the random initialized ones do. Therefore, this work proposes two variants for measuring $D(\theta_r, \theta_B)$ (illustrated in Fig. \ref{fig:Method} (a)). In our experiments, we use $\theta_{SB}$ to denote the converged parameters on B trained from scratch, and $\theta_{FB}$ to denote the fine-tuned parameters. Correspondingly, the transferabilities measured with $\theta_{SB}$ and $\theta_{FB}$ are denoted as $T_{SB}(\theta_A)$ and $T_{FB}(\theta_A)$ respectively.

\begin{table*}[]
\small
\centering
\caption{Descriptions and statistics of the datasets used in this work.}
\label{tab:datasets}
\setlength{\tabcolsep}{5mm}
\renewcommand{\arraystretch}{1.0}
\begin{tabular}{cccc}
	\toprule
	Dataset & Size (train/test) & Classes & Image description \\
	\midrule
	ImageNet & 1,281,167/50,000 & 1000 & Photos of common objects\\
	CIFAR-10 & 50,000/10,000 & 10 & Photos of common objects, image sizes $32\times 32$  \\
	CIFAR-100 & 50,000/10,000 & 100 & Photos of common objects, image sizes $32\times 32$  \\
	Caltech-101 & 3,060/6,084 & 102 & Photos/paintings/sketches of common objects \\
	CUB-200 & 5,994/5,794 & 200 & Find-grained photos of birds \\
	FGVC Aircraft & 6,667/3,333 & 100 & Find-grained photos of aircrafts\\
	Flowers & 1,088/272 & 17 & Find-grained photos of flowers \\
	UC Merced Land Use & 1,680/420 & 21 & Remote sensing images of different land uses  \\
	POCUS & 1,692/424 & 3 & Ultrasound images of COVID-19  \\
	DTD & 1,880/1,880 & 47 & Photos mainly contain textures of objects \\
	DomainNet real  & 120,906/52,041 & 345 & Photos of common objects \\
	DomainNet painting  & 50,416/21,850 & 345 & Oil Paintings, murals, drawings, tattoos \\
	DomainNet clipart  & 33,525/14,604 & 345 & Clip art images \\
	\bottomrule
\end{tabular}
\end{table*}

\paragraph{Different Parameter Scales across Layers.} Random initialization for modern DNNs considers the dimension of features in each layer an important factor. For the inputs following the standard Gaussian distribution, reasonable random initialization methods \citep{glorot2010understanding,he2015delving} make the corresponding outputs of the layer to be also standard Gaussian. This work adapts Kaiming random initialization \citep{he2015delving}. It initializes the parameters with Gaussian distribution with mean value $\mu=0$ and variance $\sigma^2=2/n_{in}$, where $n_{in}$ is the input dimension. When measuring the transferability of a DNN, the scale difference of parameters between layers causes an imbalance. The shallow layers of DNNs usually have fewer channels, which leads to larger initialized parameters and a large distance after training. Therefore, the variation of shallow layers' parameters affects $D(\theta_A,\theta_B)$ more than the deeper layers. This problem can be solved by calculating the average transferability of multiple layers (see Fig. \ref{fig:Method} (b)):
\begin{equation}
    T_B(\theta_A) = \frac{1}{L} \sum_{l=1}^L T_B(\theta_A^{(l)}),
\end{equation}
where $\theta_A^{(l)}$ and $L$ denote the parameters of layer $l$ and the total number of layers, respectively. It is worth noticing that, the transferability of layer $l$, $T_B(\theta_A^{(l)})$, can also be used to measure the trend of transferability between layers.

\section{Experiments and Results}

This section describes our experimental results in detail. We begin with the datasets and metrics to evaluate domain gaps and widths. Then we describe our implementation of transfer learning. Finally, we demonstrate and discuss the transferabilities of pre-trained parameters in different scenarios.

\subsection{Datasets and Metrics}

Our experiments include transfer learning from ImageNet to twelve image classification datasets (CIFAR-10/CIFAR-100 \citep{krizhevsky2009learning}, Caltech-101 \citep{fei2004learning}, CUB-200 \citep{wah2011caltech}, FGVC Aircraft \citep{maji2013fine}, Flowers \citep{nilsback2006visual}, UC Merced Land Use \citep{yang2010bag}, POCUS \citep{born2020pocovid}, DTD \citep{sharan2014accuracy}, DomainNet \citep{peng2019moment}) from a wide range. This is because we find the experiments conducted by previous works \citep{yosinski2014transferable,azizpour2015factors} are primarily on natural images (\ie photos), missing the images from farther domains (\eg medical images, remote sensing images, paintings) or wider domains (\eg DomainNet). But in many transfer learning applications, the distribution of the target domain is often significantly different from the source domain. The brief information of the datasets used in this work is summarized in Tab.\ref{tab:datasets}

To quantify the domain gaps of datasets to ImageNet as well as the domain widths, previous methods used trained DNNs to extract semantic features of different datasets \citep{liu2022closer,stacke2020measuring}. An obvious drawback is that DNNs trained on specific tasks would have skewed performance on other tasks. In this work, we need the domain metrics to be domain-agnostic so that the gaps between different datasets can be more meaningful and comparable. The basic image color and texture features are desirable for evaluating the feature distributions of different datasets because they are not related to the categorical semantics~\citep{saito2019strong}.
First, the RGB channels' mean value and standard deviation are considered the color features. Second, \textit{Gray-Level Co-occurrence Matrix} features~\citep{haralick1973textural} (Angular Second Moment, Entropy, Contrast, and Inverse Differential Moment) of 4 directions are adopted as the texture features. The total feature dimension is $3 + 3 + 4\times 4 = 22$. 
Given feature distributions, the domain gap is measured as the \textit{Maximum Mean Discrepancy} \citep{gretton2012kernel} between two datasets. Finally, the domain width is calculated as the maximum eigenvalue of the covariance matrix of features in a dataset. The domain gaps and widths of datasets used in this work are illustrated in Fig. \ref{fig:domain}, which is highly consistent with the common expectation (see supplementary material for the illustration of all datasets included in this work).

\begin{figure}[t]
\centering\centerline{\includegraphics[width=1.0\linewidth]{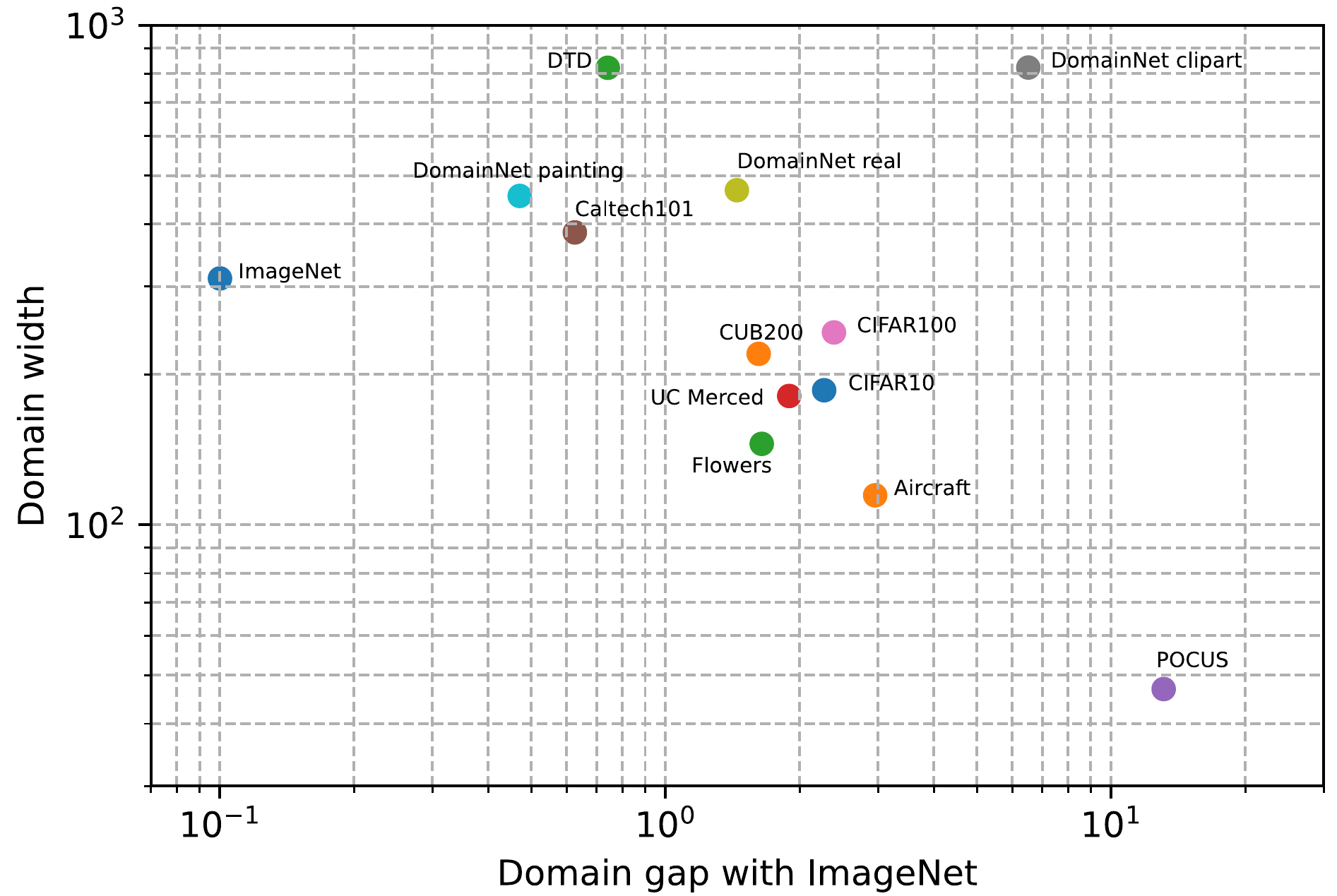}}
\caption{The domain gaps with ImageNet and the domain widths of downstream datasets. For good illustration, the domain gap between ImageNet and itself is set to 0.1.}
\label{fig:domain}
\end{figure}

\begin{table*}[]
\tiny
\centering
\caption{Dataset information (\ie the domain gap compared with ImageNet, domain width, and training data amount), training performance (random initialization vs. fine-tuning), and transferabilities on different downstream tasks.}
\label{tab:exp1}
\setlength{\tabcolsep}{2mm}  % width
\renewcommand{\arraystretch}{1.0}  % height
\begin{tabular}{ccccccccccccc}
	\toprule
	   & CIFAR-10 & CIFAR-100 & Caltech-101 & CUB-200 & Aircraft & Flowers & Land Use & POCUS & DTD & DomainNet-r & DomainNet-p & DomainNet-c \\
	\midrule
	  Domain gap  & 2.27 & 2.39 & 0.63 & 1.62 & 2.95 & 1.64 & 1.89 & 13.12 & 0.74 & 1.45 & 0.47 & 6.56  \\
	  Domain width & 186.0 & 242.9 & 385.0 & 220.0 & 114.6 & 145.3 & 181.2 & 46.9 & 822.3 & 467.7 & 455.5 & 823.2 \\
	  Data amount & 50k & 50k & 3.1k & 6.0k & 6.7k & 1.1k & 1.7k & 1.7k & 1.9k & 120k & 50k & 34k \\
	  \midrule
	  Random init (\%) & 84.9 & 54.6 & 64.6 & 32.4 & 41.5 & 72.8 & 83.6 & 85.9 & 29.3 & 71.4 & 48.5 & 53.5  \\
	  Fine-tuning (\%) & 90.0 & 66.3 & 94.6 & 76.4 & 81.6 & 98.9 & 98.8 & 94.6 & 73.1 & 81.7 & 69.0 & 74.9 \\
	  Relative error rate (\%) $\downarrow$ & 33.8 & 25.8 & 84.7 & 65.1 & 68.5 & 96.0 & 92.7 & 61.7 & 62.0 & 36.0 & 39.8 & 46.0 \\
	  \midrule
	  $T_{FB}(\theta_A)$ & 7.79 & 5.59 & 23.92 & 21.96 & 12.45 & 106.30 & 39.09 & 53.58 & 24.15 & 6.32 & 7.12 & 8.91 \\
	  $T_{SB}(\theta_A)$ & 2.79 & 1.93 & 3.15 & 2.61 & 1.83 & 6.94 & 3.15 & 2.73 & 2.64 & 3.17 & 2.30 & 2.32 \\
	\bottomrule
\end{tabular}
\end{table*}

\subsection{Implementation Details}

We utilize the ImageNet dataset to pre-train ResNet-50 \citep{he2016deep} backbone (experiments with other backbones are demonstrated in the supplementary material, which shows similar conclusions). The hyperparameters are set to be the same as the original paper. For fine-tuning, we use $32\times 32$ CIFAR images and the input sizes of the other datasets are set to $224\times 224$. Batch size of 128, and training time of 100 epochs are used for all datasets. We use the initial learning rate (LR) of 0.01, SGD optimizer with the momentum of 0.9, and weight decay of 0.0001. The SGD optimizer is crucial for keeping learning rates for different layers consistent (adaptive optimizers control LRs in an unexpected way). The LRs decrease with a cosine LR scheduler. All images are augmented with random cropping and horizontal flipping with the probability of 0.5. When training DNNs from scratch, all hyperparameters are set to be identical as fine-tuning for a fair comparison. For all experiments with random initialization, initial parameters are also kept the same.

When measuring the transferabilities of the whole DNN, the last fully connected layer will be excluded because it is meaningless to transfer parameters that only work for specific tasks.

\subsection{Verification of the Transferability Measurement}

In Tab. \ref{tab:exp1}, the information of datasets, training performance from random initialization and fine-tuning, and the transferabilities on different tasks are demonstrated. The quantified transferabilities, $T_{FB}(\theta_A)$ and $T_{SB}(\theta_A)$, are highly correlated with the error rate decreasing ratio (p = 0.0001 and 0.0002 in F-test, respectively). This result indicates the rationality of the proposed transferability measurement because, with a higher transferability, a DNN can improve the downstream performance to a larger extent. Compared with the downstream performance, such as accuracy, the transferability defined in this work is less limited because it depends only on parameter variation, which is model-agnostic and task-agnostic.

\subsection{Factors Affecting Transferability}

In previous works, researchers usually regard DNNs to be less transferable when domain gaps are larger \citep{yosinski2014transferable,azizpour2015factors,tajbakhsh2016convolutional}. We find that this law does hold generally, and transferability is even more negatively affected by the domain width and data amount of downstream tasks. 

In Tab. \ref{tab:exp1}, datasets having larger domain gaps with ImageNet tend to have lower transferabilities. For example, CIFAR-10 and CIFAR-100 have similar domain widths and data amounts, but CIFAR-100 has a larger domain gap than CIFAR-10, leading to lower transferability. The same phenomenon can also be observed between the Flowers and Land Use datasets. However, we can also see that CUB-200 and Flowers have similar domain gaps with ImageNet, but pre-trained DNNs have different transferabilities. This fact motivates our interest in exploring other crucial factors for transferability. Domain width and data amount are chosen as possible factors, as discussed in the introduction.

\begin{figure}[t]
\centering\centerline{\includegraphics[width=1.0\linewidth]{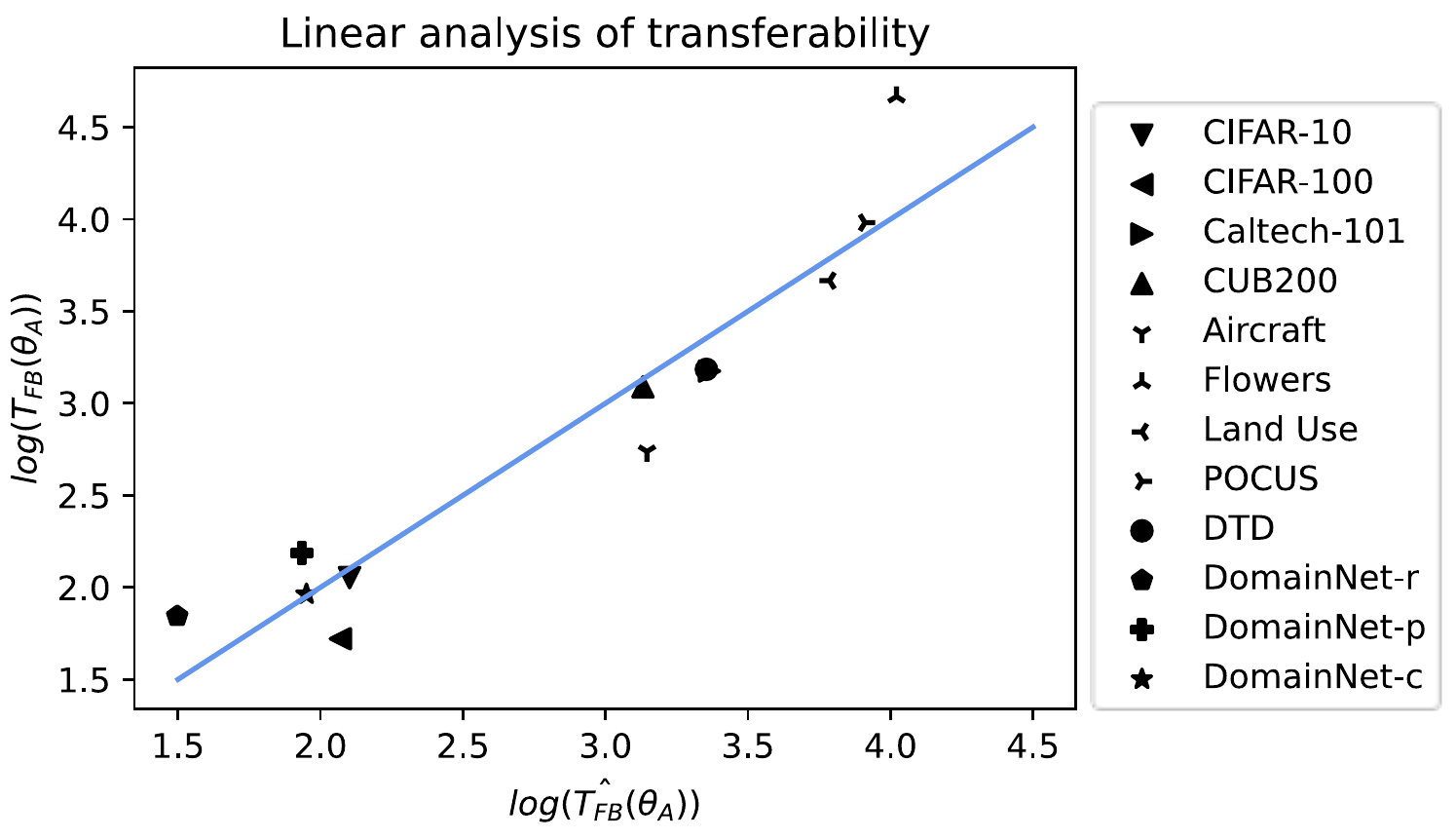}}
\caption{Linear analysis of different factors. The correlation coefficient is 0.90.}
\label{fig:linear}
\end{figure}

In this work, we validate the three factors by linear analysis. By fitting $log(T_{B}(\theta_A))$ to the domain gap $\mathcal{G}$, domain width $\mathcal{W}$, and log data amount $log(\mathcal{N})$, we find a high coefficient of determination for $log(T_{FB}(\theta_A))$ ($r^2$=0.90, see Fig. \ref{fig:linear}). The fitted linear relationship is

\begin{equation}
    log(\hat{T}_{FB}(\theta_A)) = -0.02 \mathcal{G} - 0.14 \mathcal{W} - 0.80 log(\mathcal{N}) + 2.86.
\label{eq:linear}
\end{equation}

We can see that, despite the slightly negative effect of the domain gap, there are stronger negative correlations between transferabilities and domain width and data amount. The reason will be elaborated on later. For $log(T_{SB}(\theta_A))$, linear regression performs poorly ($r^2 = 0.26$), indicating that the three factors can hardly interpret $T_{SB}(\theta_A)$. This result may be due to the unstable convergence points of random initialized DNNs. Some convergence points are too bad to be considered in the ``solution set" of the tasks (\eg CUB and DTD). Therefore, $D(\theta_r,\theta_{SB})$ is not a desirable metric for measuring transferability. $T_{FB}(\theta_A)$ is mainly used for the following analysis.

\begin{figure}[t]
\centering\centerline{\includegraphics[width=1.0\linewidth]{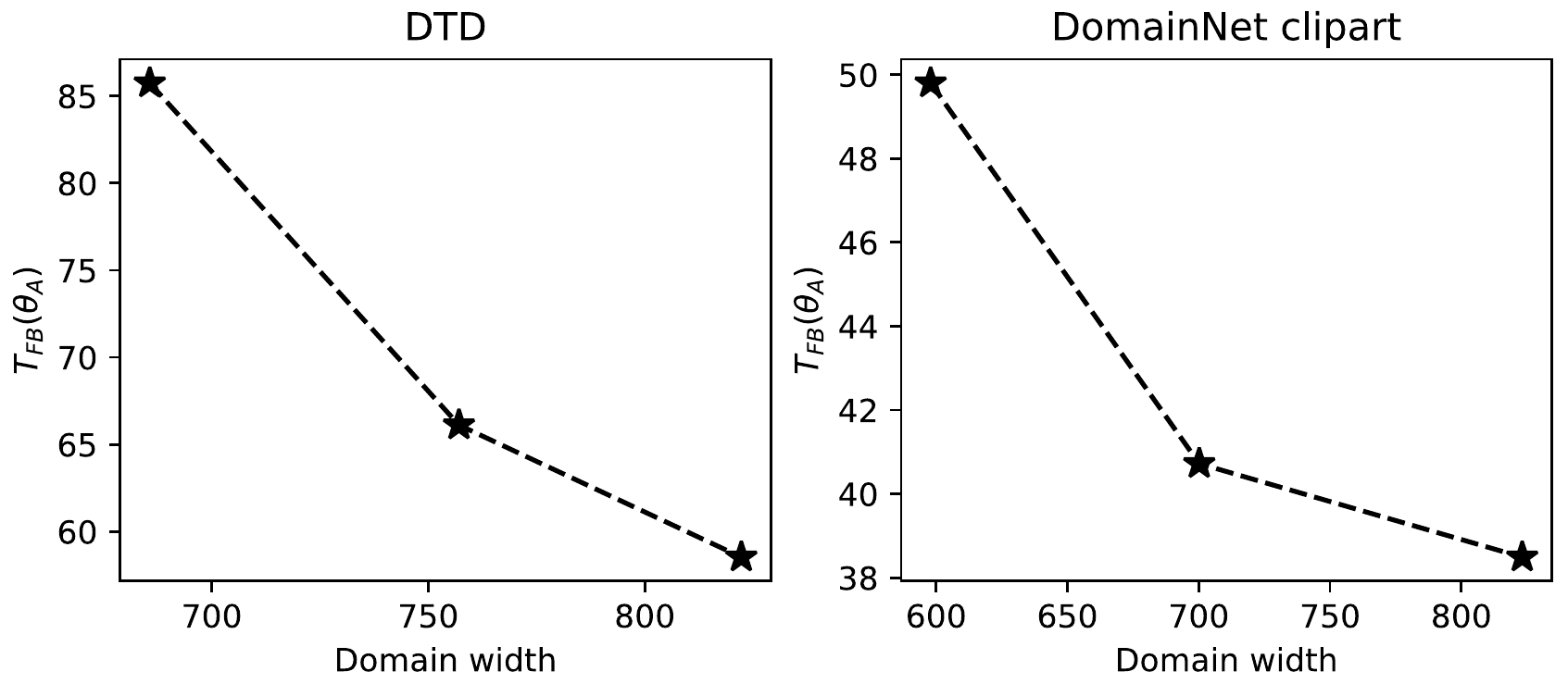}}
\caption{The negative correlation between domain widths and transferabilities.}
\label{fig:width}
\end{figure}

\paragraph{Domain Width.}
% decreasing classes of DTD and DomainNet-c while data amounts keep the same
We try to remove distractions from domain gap and data amount, and explore the relationship between domain width and transferability $T_{FB}(\theta_A)$. DTD and DomainNet-c are selected as the target datasets as they have the largest domain width. For DTD, we keep the training data amount to 1000 and use 12, 20, and 47 classes to calculate the transferabilities. It is worth noting that the domain gaps with ImageNet change slightly (from 6.56 to 6.44) when the number of classes decreases. For DomainNet-c, the domain gap is also stable when we randomly sample 2,000 images from 35, 100, or 345 classes. The results are shown in Fig. \ref{fig:width}, where the transferabilities decrease as the domain widths increase, which validates Eqn. (\ref{eq:linear}). 

\begin{figure}[t]
\centering\centerline{\includegraphics[width=1.0\linewidth]{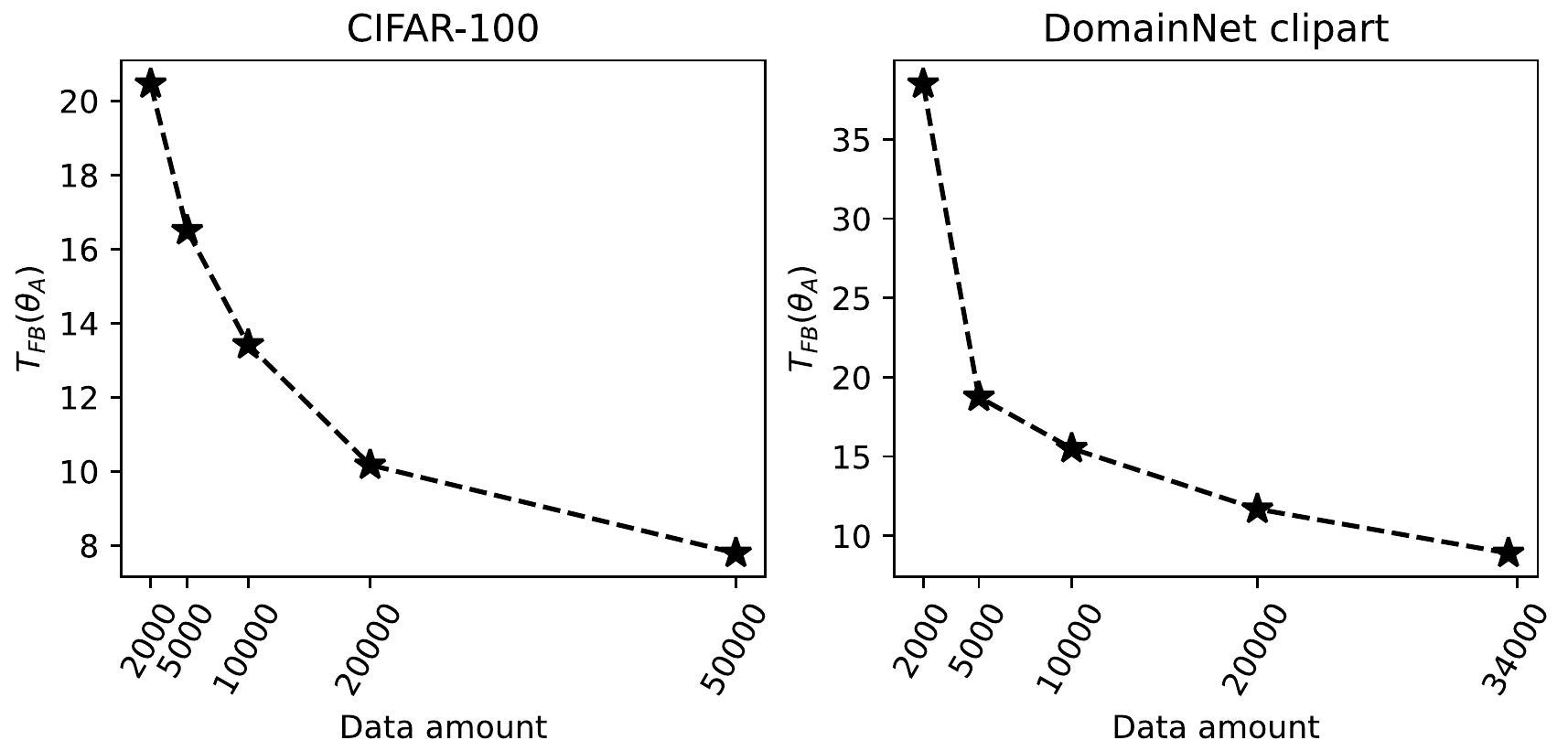}}
\caption{The negative correlation between downstream data amounts and transferabilities.}
\label{fig:exp2}
\end{figure}

\paragraph{Downstream Data Amount.}
This experiment verifies the negative correlation between transferabilities and data amounts. We uniformly sample data from all classes of CIFAR-100 and DomainNet-c with different amounts, and the transferability curves are shown in Fig. \ref{fig:exp2}. It is worth mentioning that the random sampling scheme makes the domain gap and domain width stable in this experiment. With the increasing amount of data, the transferabilities decrease monotonically. Moreover, comparing Fig. \ref{fig:width} with Fig. \ref{fig:exp2}, the effect of data amount seems greater than that of domain width, which is consistent with the Eqn. (\ref{eq:linear}).

\begin{figure}[t]
\centering\centerline{\includegraphics[width=0.8\linewidth]{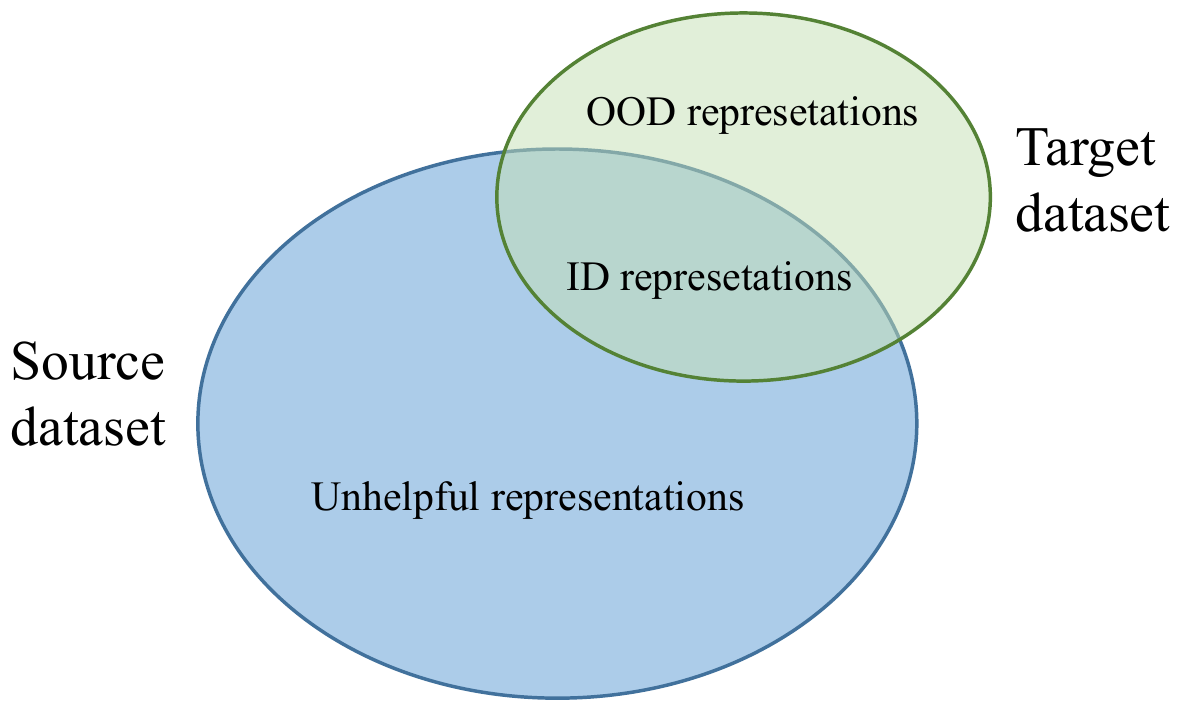}}
\caption{A conceptual visualization of source dataset and target dataset. ``OOD" and ``ID" are the abbreviations for ``out-of-distribution" and ``in-distribution".}
\label{fig:explanation}
\end{figure}

\paragraph{A General Explanation.}
We try to explain the negative effects of domain gap, domain width, and data amount with a general framework (Fig. \ref{fig:explanation}). In the pre-training phase, a DNN learns the basic knowledge (also called visual representations for visual tasks) from the source dataset. Within these representations, there are some in-distribution representations that can be used for the downstream task directly. However, there are still new visual representations that the DNN needs to learn from the downstream task for better performance. In the fine-tuning phase, the pre-trained DNN updates its parameters to learn these out-of-distribution representations. Such a larger parameter change corresponds to lower transferability. When the domain gap becomes greater or the domain width becomes larger, the more out-of-distribution representations the DNN needs to learn, and the lower transferability the DNN has. Similarly, given the fixed domain gap and domain width, higher downstream data amount makes the representations that need to be learned denser. The wider range and denser new representations decrease the DNN transferability by increasing knowledge that cannot be transferred from experiences.

\subsection{Layer-wise Transferability}\label{sect:layer_trans}

\begin{figure}[t]
\centering\centerline{\includegraphics[width=1.0\linewidth]{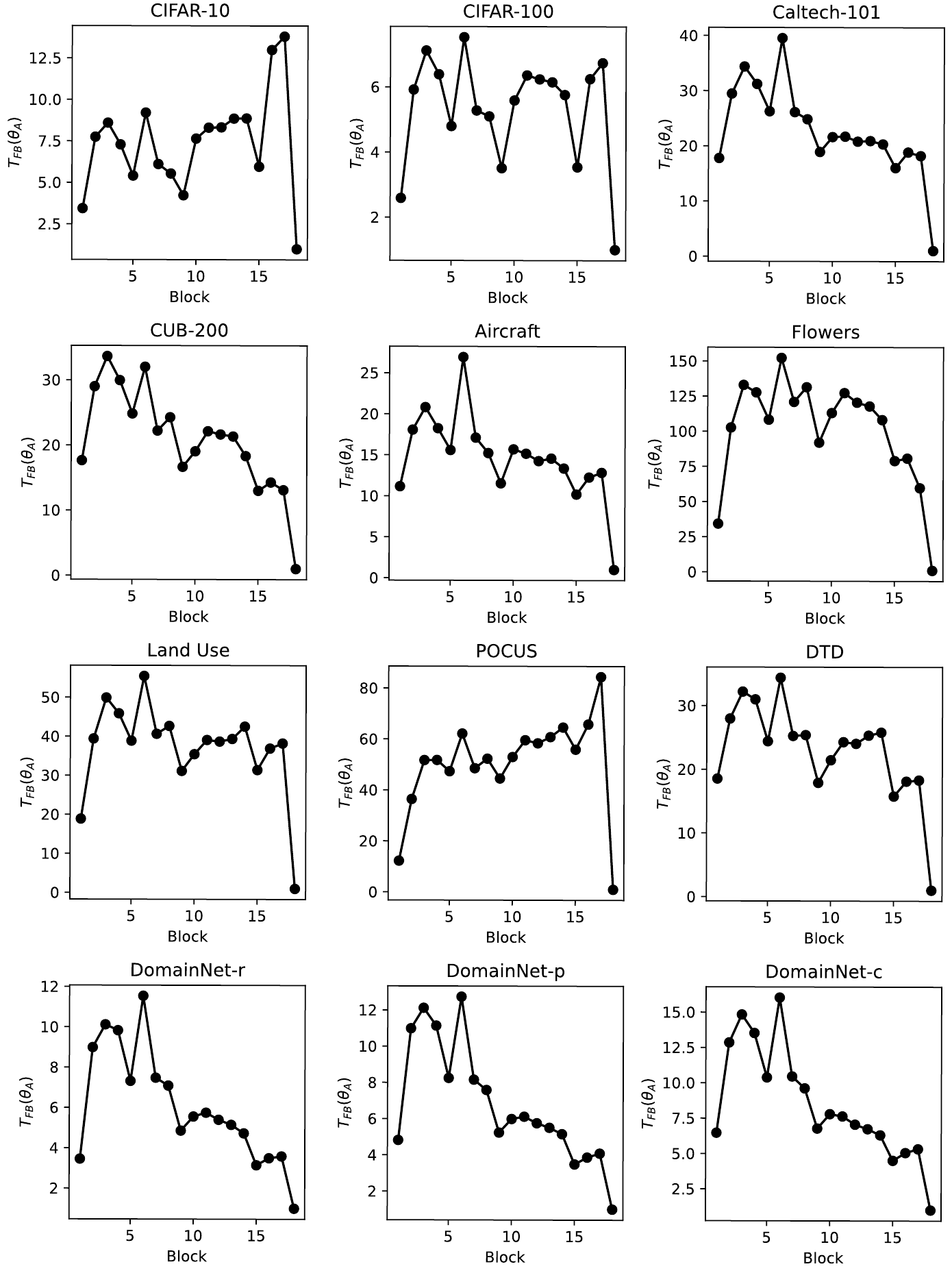}}
\caption{The layer-wise transferabilities on different datasets. The ``Block" denotes the first convolutional layer, the bottle-neck (a three-conv-layer residual block), or the last fully connected layer of ResNet.}
\label{fig:layerwise}
\end{figure}

Previous literature usually regards the lower layers to be more transferable \citep{yosinski2014transferable}, due to the generality of the low-level feature extraction abilities. This work shows that this is not always the case. The trend of transferabilities across layers is illustrated in Fig. \ref{fig:layerwise}. There are two noteworthy laws. First, nearly all layer-wise transferabilities are inverted U-shaped instead of downward trends, mainly caused by the first two poorly transferable layers. Second, discarding the first two layers, nine datasets have descending layer-wise transferabilities across layers consistent with our previous consensus. However, three datasets (CIFAR-10, CIFAR-100, and POCUS) still do not present any meaningful upward/downward trend. For these three datasets, we find that the low-level features may not be transferable as the low-level image pattern are not similar anymore. For CIFAR-10 and CIFAR-100, the image sizes are only $32\times 32$. The pre-trained ResNets have learned visual representations from $224\times 224$ ImageNet data, so these representations may be poorly transferable to the seriously different local pattern distributions. For the POCUS dataset, the ultrasound images also have quite different image patterns compared with natural images, making the pre-trained layers for extracting local features harder to transfer. All in all, the different pattern distributions of pre-training and downstream datasets make the layer-wise transferability no longer decreasing.

\begin{figure}[t]
\centering\centerline{\includegraphics[width=1.0\linewidth]{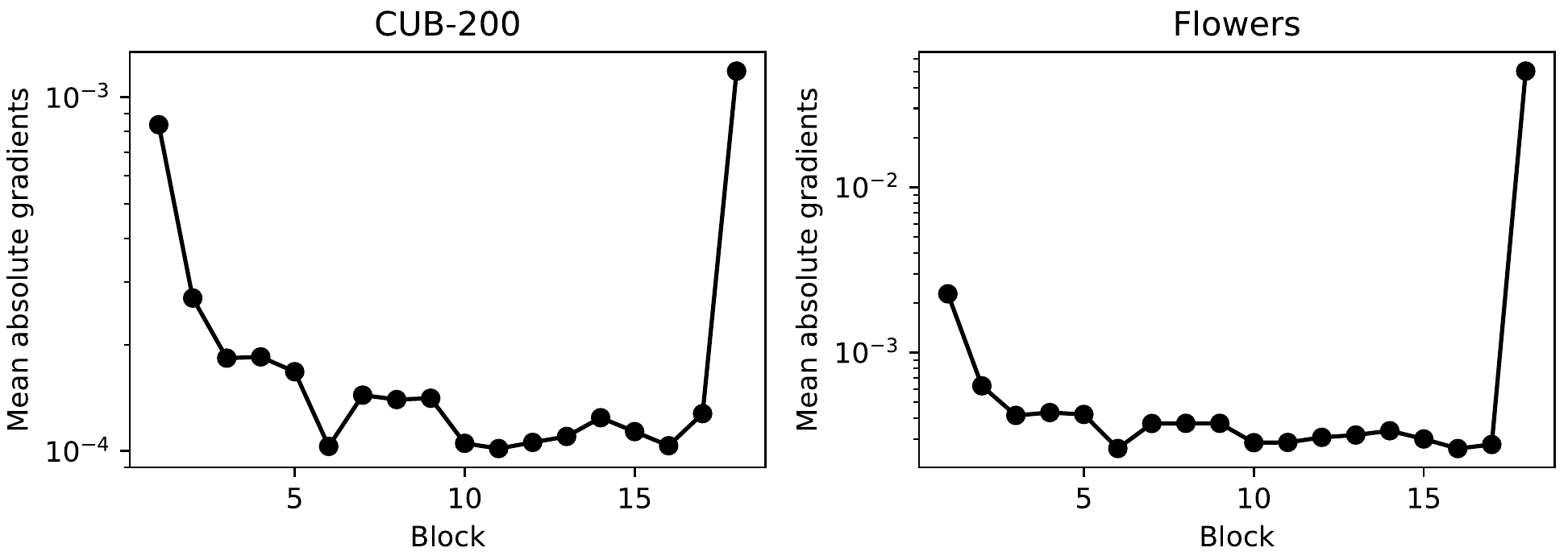}}
\caption{The layer-wise mean absolute gradients on CUB and Flower datasets. Both curves are U-shaped.}
\label{fig:layerwise_grad}
\end{figure}

\paragraph{Domain-sensitivity of the Lowest Layers.}
% The gradient of each layer of pre-trained DNNs on downstream tasks.
In Fig. \ref{fig:layerwise}, we can see a common phenomenon: the first layer has very low transferability on all datasets, including those natural image datasets with small domain gaps. This result may indicate that the first layer is more sensitive to the domain shift from ImageNet than the subsequent layers. We assume that when a pre-trained DNN ``sees” a new sample with out-of-distribution basic visual patterns, the first layer weights would get large gradients and update drastically. These kinds of out-of-distribution basic visual patterns may be common in downstream datasets. To verify this assumption, we calculate the layer-wise gradients of pre-trained DNNs on two natural image datasets, \ie CUB-100 and Flowers (see Fig. \ref{fig:layerwise_grad}), and we find that both curves are U-shaped. It is reasonable for the last layers to have large gradients, as they are replaced and randomly initialized for specific tasks. But for the large gradients of the pre-trained lower layers, they suggest that the parameters of these layers are somehow ``unfit"  for the current tasks, which validates our assumption.

\begin{figure}[t]
\centering\centerline{\includegraphics[width=1.0\linewidth]{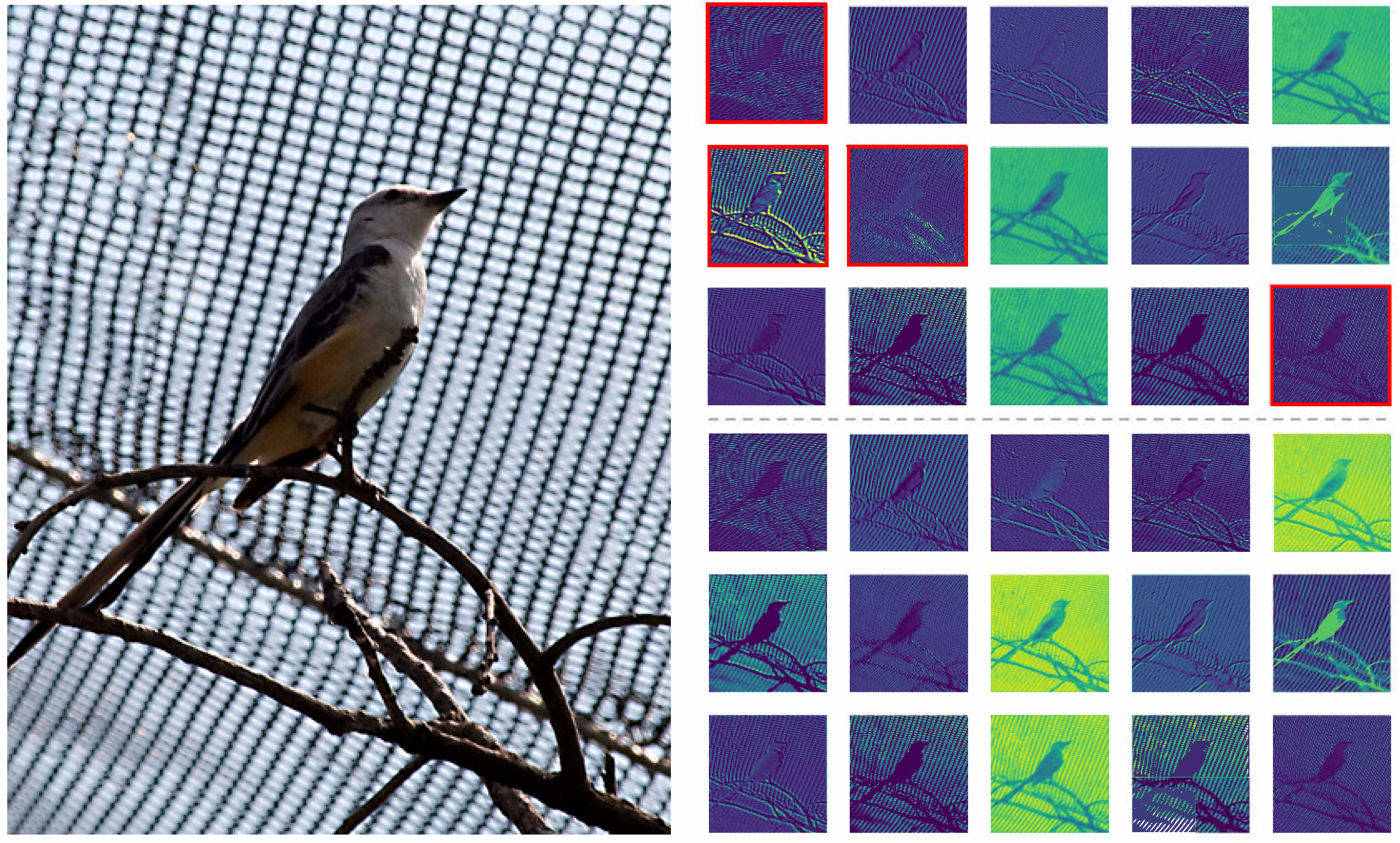}}
\caption{Input image (left), first-layer features of pre-trained DNN (upper right), and the corresponding features after fine-tuning (lower right). For feature maps, lighter means more activated.}
\label{fig:feature_maps}
\end{figure}

\paragraph{Visualization of Pre-trained/Fine-tuned Features.}
% The feature maps before/after fine-tuning.
To further illustrate the weaker transferabilities of lower layers, we choose CUB-200 to show how first-layer features of a pre-trained DNN change after fine-tuning. In Fig. \ref{fig:feature_maps}, the pre-trained first layer can successfully activate meaningful edges. But some features are not good enough (in red boxes), with unclear edges or activating the background gauze too much. After fine-tuning, these imperfect feature maps become clearer and more helpful for recognizing the bird. 
As reported in \citep{donahue2014decaf}, a unit's activation is better be maximized to extract more meaningful image features, meaning that the fine-tuned parameters are more capable of finding key patterns on the images. The benefit of fine-tuning the first layer is increasing the activations of the target object in the image (mean value increases from 0.268 to 0.284). The activation distributions of pre-trained/fine-tuned DNNs significantly differ with the $t$ statistic of -16.99, $p=4.6\times 10^{-57}$, which shows significant improvement brought by updating the first layer. This result indicates that some previous methods of only tuning higher layers~\citep{tajbakhsh2016convolutional} can still be improved concerning the first-layer transferability.

\section{Conclusion}

In this work, we have proposed a novel method for quantifying the transferabilities of DNNs. We conducted extensive experiments on twelve datasets with this method and showed how transferability is more negatively affected by domain width and downstream data amount than domain gap. This phenomenon suggests that downstream tasks with narrower domains and fewer data benefit more from pre-training. Transfer learning with a large domain gap (\eg medical data) may reach higher performance considering this result. In addition, we observed that the layer-wise transferabilities of DNNs are not monotonically decreasing as the traditional consensus regarded. The root cause is that the first layers of DNNs are sensitive to domains. They have larger gradients than the middle layers and are usually imperfect for downstream tasks. Fine-tuning techniques with the least transferable layers from both ends may achieve better performance than previous works with only the last layers.

\bibliography{aaai22}
\end{document}